\documentclass[nonatbib]{article}

\usepackage[final]{nips_2016}

\usepackage[utf8]{inputenc} 
\usepackage[T1]{fontenc}    
\usepackage{url}            
\usepackage{booktabs}       
\usepackage{amsfonts}       
\usepackage{nicefrac}       
\usepackage{microtype}      

\usepackage{amsmath}
\usepackage{amsthm}
\usepackage{amsfonts}

\usepackage{floatrow}

\usepackage{graphicx}
\usepackage{wrapfig}

\usepackage[numbers]{natbib}

\theoremstyle{theorem}
\newtheorem{thrm}{Theorem}

\theoremstyle{theorem}

\usepackage{hyperref}       

\newcommand{\mb}[1]{\mathbf{#1}}
\newcommand{\bs}[1]{\boldsymbol{#1}}

\title{Improving Variational Auto-Encoders\\using Householder Flow}

\author{
  Jakub M. Tomczak, Max Welling \\
  University of Amsterdam\\
  \texttt{J.M.Tomczak@uva.nl, M.Welling@uva.nl}
  }

\begin{document}

\maketitle

\begin{abstract}
Variational auto-encoders (VAE) are scalable and powerful generative models. However, the choice of the variational posterior determines tractability and flexibility of the VAE. Commonly, latent variables are modeled using the normal distribution with a diagonal covariance matrix. This results in computational efficiency but typically it is not flexible enough to match the true posterior distribution. One fashion of enriching the variational posterior distribution is application of \textit{normalizing flows}, \textit{i.e.}, a series of invertible transformations to latent variables with a simple posterior. In this paper, we follow this line of thinking and propose a \textit{volume-preserving flow} that uses a series of \textit{Householder transformations}. We show empirically on MNIST dataset and histopathology data that the proposed flow allows to obtain more flexible variational posterior and competitive results comparing to other normalizing flows.
\end{abstract}


\section{Variational Auto-Encoder}

Let $\mb{x}$ be a vector of $D$ observable variables, $\mb{z} \in \mathbb{R}^{M}$ a vector of stochastic latent units (variables) and let $p(\mb{x}, \mb{z})$ be a parametric model of the joint distribution. Given $N$ datapoints $\mb{X} = \{\mb{x}_1, \ldots, \mb{x}_N\}$ we typically aim at maximizing the marginal log-likelihood:
\begin{equation}\label{eq:loglikelihood}
\ln p(\mb{X}) = \sum_{i=1}^{N} \ln p(\mb{x}_{i}),
\end{equation}
with respect to parameters. This task could be troublesome due to intractability of the marginal likelihood, \textit{e.g.}, when the model is parameterized by a neural network (NN). To overcome this issue one can introduce an \textit{inference model} (an \textit{encoder}) $q(\mb{z}|\mb{x})$ and optimize the variational lower bound:
\begin{equation}\label{eq:elbo}
\ln p(\mb{x}) \geq \mathbb{E}_{q(\mb{z}|\mb{x})}[ \ln p(\mb{x}|\mb{z}) ] - \mathrm{KL} \big{(} q(\mb{z}|\mb{x}) || p(\mb{z}) \big{)},
\end{equation}
where $p(\mb{x}|\mb{z})$ is called a \textit{decoder} and $p(\mb{z}) = \mathcal{N}(\mb{z}|\mb{0}, \mb{I})$ is the \textit{prior}. There are various ways of optimizing this lower bound but for continuous $\mb{z}$ this could be done efficiently through a re-parameterization of $q(\mb{z}|\mb{x})$ \citep{KW:13, RMW:14}. Then the architecture is called a \textit{variational auto-encoder} (VAE).

In practice, the inference model assumes a diagonal covariance matrix, \textit{i.e.}, $q(\mb{z}|\mb{x}) = \mathcal{N}\big{(} \mb{z}|\boldsymbol\mu (\mb{x}), \mathrm{diag}(\boldsymbol\sigma ^{2}(\mb{x})) \big{)}$, where $\boldsymbol\mu (\mb{x})$ and $\boldsymbol\sigma ^{2}(\mb{x})$ are parameterized by the NN. However, the assumption about the diagonal posterior can be insufficient and not flexible enough to match the true posterior.

\section{Improving posterior flexibility using \textit{Normalizing Flows}}
A (finite) \textit{normalizing flow}, first formulated by \citep{TT:13, TV:10} and further developed by \citep{RM:15}, is a powerful framework for building flexible posterior distribution by starting with an initial random variable with a simple distribution for generating $\mb{z}^{(0)}$ and then applying a series of invertible transformations $\mb{f}^{(t)}$, for $t=1,\ldots , T$. As a result, the last iterate gives a random variable $\mb{z}^{(T)}$ that has a more flexible distribution. Once we choose transformations $\mb{f}^{(t)}$ for which the Jacobian-determinant can be computed, we aim at optimizing the following objective:
\begin{equation}\label{eq:nfobjective}
\ln p(\mb{x}) \geq \mathbb{E}_{q(\mb{z}^{(0)}|\mb{x})} \Big{[} \ln p(\mb{x}|\mb{z}^{(T)}) + \sum_{t=1}^{T} \ln \Big{|}\mathrm{det}\frac{\partial \mb{f}^{(t)} }{ \partial \mb{z}^{(t-1)} } \Big{|} \Big{]} - \mathrm{KL} \big{(} q(\mb{z}^{(0)}|\mb{x}) || p(\mb{z}^{(T)}) \big{)}.
\end{equation}
In fact, the normalizing flow can be used to enrich the posterior of the VAE with small or even none modifications in the architecture of the encoder and the decoder. 

There are two main kinds of normalizing flows, namely, \textit{general normalizing flows} and \textit{volume preserving flows}. The difference between these types of flow is in the manner how the Jacobian-determinant is handled. The general normalizing flows aim at formulating the flow for which the Jacobian-determinant is relatively easy to compute. On the contrary, the volume-preserving flows design series of transformations such that the Jacobian-determinant equals $1$ while still it allows to obtain flexible posterior distributions. The reduced computational complexity is the main reason why volume-preserving flows are so appealing. The question is whether one can propose a series of transformations for which the Jacobian-determinant is equal one, they are cheap to calculate and are general enough to model flexible posteriors. In the next subsection we present a new volume-preserving flow that applies series of Householder transformations that we refer to as the \textit{Householder flow}.

\section{Householder Flow}

\subsection{Motivation}

In general, any full-covariance matrix $\bs{\Sigma}$ can be represented by the eigenvalue decomposition using eigenvectors and eigenvalues:
\begin{equation}\label{eq:eigenvalue_decomposition}
\bs{\Sigma} = \mb{U} \mb{D} \mb{U}^{\top} , 
\end{equation}
where $\mb{U}$ is an orthogonal matrix with eigenvectors in columns, $\mb{D}$ is a diagonal matrix with eigenvalues. In the case of the vanilla VAE, it would be tempting to model the matrix $\mb{U}$ to obtain a full-covariance matrix. The procedure would require a linear transformation of a random variable using an orthogonal matrix $\mb{U}$. Since the absolute value of the Jacobian determinant of an orthogonal matrix is $1$, for $\mb{z}^{(1)} = \mb{U} \mb{z}^{(0)}$ one gets $\mb{z}^{(1)} \sim \mathcal{N}( \mb{U}\bs{\mu}, \mb{U}\  \mathrm{diag}(\bs{\sigma}^{2})\ \mb{U}^{\top} )$. If $\mathrm{diag}(\bs{\sigma}^{2})$ coincides with true $\mb{D}$, then it would be possible to resemble the true full-covariance function. Hence, the main goal would be to model the orthogonal matrix of eigenvectors.

Generally, the task of modelling an orthogonal matrix in a principled manner is rather non-trivial. However, first we notice that any orthogonal matrix can be represented in the following form \citep{BS:94, SB:95}:
\begin{thrm}\label{theorem:kernel_representation} \emph{(The Basis-Kernel Representation of Orthogonal Matrices)}\\
For any $M \times M$ orthogonal matrix $\mb{U}$ there exist a full-rank $M \times K$ matrix $\mb{Y}$ (the \textit{basis}) and a nonsingular (triangular) $K \times K$ matrix $\mb{S}$ (the \textit{kernel}), $K \leq M$, such that:
\begin{equation}
\mb{U} = \mb{I} - \mb{Y} \mb{S} \mb{Y}^{\top} .
\end{equation}
\end{thrm}
The value $K$ is called the \textit{degree} of the orthogonal matrix. Further, it can be shown that any orthogonal matrix of degree $K$ can be expressed using the product of Householder transformations \citep{BS:94, SB:95}, namely:
\begin{thrm}\label{theorem:number_of_householders}
Any orthogonal matrix with the basis acting on the $K$-dimensional subspace can be expressed as a product of exactly $K$ Householder transformations:
\begin{equation}
\mb{U} = \mb{H}_{K} \mb{H}_{K-1} \cdots \mb{H}_{1},
\end{equation}
where $\mb{H}_{k} = \mb{I} - \mb{S}_{kk} \mb{Y}_{\cdot k} (\mb{Y}_{\cdot k})^{\top}$, for $k=1,\ldots,K$.
\end{thrm}

Theoretically, Theorem \ref{theorem:number_of_householders} shows that we can model any orthogonal matrix in a principled fashion using $K$ Householder transformations. Moreover, the Householder matrix $\mb{H}_{k}$ is \textit{orthogonal} matrix itself \citep{H:58}. Therefore, this property and the Theorem \ref{theorem:number_of_householders} put the Householder transformation as a perfect candidate for formulating a volume-preserving flow that allows to approximate (or even capture) the true full-covariance matrix.

\subsection{Definition}

The \textit{Householder transformation} is defined as follows. For a given vector $\mb{z}^{(t-1)}$ the reflection hyperplane can be defined by a vector (a \textit{Householder vector}) $\mb{v}_{t} \in \mathbb{R}^{M}$ that is orthogonal to the hyperplane, and the reflection of this point about the hyperplane is \citep{H:58}:
\begin{align}\label{eq:reflection}
\mb{z}^{(t)} &= \Big{(} \mb{I} - 2 \frac{\mb{v}_{t}\mb{v}_{t}^{\top}}{||\mb{v}_{t}||^{2}} \Big{)} \mb{z}^{(t-1)} \\
&= \mb{H}_{t} \mb{z}^{(t-1)},
\end{align}
where $\mb{H}_{t} = \mb{I} - 2 \frac{\mb{v}_{t}\mb{v}_{t}^{\top}}{||\mb{v}_{t}||^{2}}$ is called the \textit{Householder matrix}.

\begin{figure}[!htbp]
\includegraphics[width=1.0\textwidth]{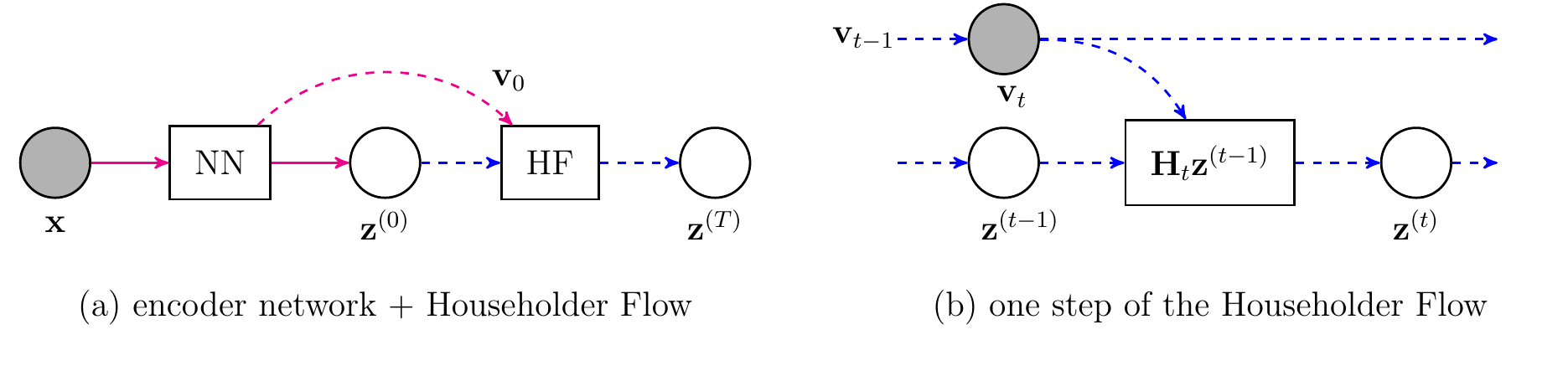}
\vskip -0.5cm
\caption{A schematical representation of the encoder network with the Householder flow. (a) The general architecture of the VAE+HF: The encoder returns means and variances for the posterior and the first Householder vector that is further used to formulate the Householder flow. (b) A single step of the Householder flow that uses linear Householder transformation. In both figures solid lines correspond to the encoder network and the dashed lines are additional quantities required by the HF. }
\label{fig:householder_flow}
\end{figure}

The most important property of $\mb{H}_{t}$ is that it is an orthogonal matrix and hence the absolute value of the Jacobian determinant is equal $1$. This fact significantly simplifies the objective (\ref{eq:nfobjective}) because $\ln \Big{|}\mathrm{det}\frac{\partial \mb{H}_{t}\mb{z}^{(t-1)} }{ \partial \mb{z}^{(t-1)} } \Big{|} = 0$, for $t=1,\ldots,T$. Starting from a simple posterior with the diagonal covariance matrix for $\mb{z}^{(0)}$, the series of $T$ linear transformations given by (\ref{eq:reflection}) defines a new type of volume-preserving flow that we refer to as the \textit{Householder flow} (HF). The vectors $\mb{v}_{t}$, $t=1,\ldots, T$, are produced by the encoder network along with means and variances using a linear layer with the input $\mb{v}_{t-1}$, where $\mb{v}_{0} = \mb{h}$ is the last hidden layer of the encoder network. The idea of the Householder flow is schematically presented in Figure \ref{fig:householder_flow}. Once the encoder returns the first Householder vector, the Householder flow requires $T$ linear operations to produce a sample from a more flexible posterior with an approximate full-covariance matrix.

\section{Related Work}

In \citep{RM:15} invertible linear normalizing flows with known Jacobian determinant were proposed. These are easy to calculate, \textit{i.e.}, the determinant of the Jacobian can be analytically computed, however, many such transformations are needed to capture high-dimensional dependencies. A different approach relies on \textit{volume-preserving flows}, for which the absolute value of Jacobian determinant is equal $1$, such as, Non-linear Independent Components Estimation (NICE) \citep{DKB:14}, Hamiltonian Variational Inference (HVI) \citep{SKW:15} and linear Inverse Autoregressive Flow (linIAF) or its non-linear version (nlIAF) \citep{KSJCSW:16}. 

The HF is similar in spirit to the linIAF where the linear transformation is also applied but it is the lower triangular inverse Cholesky matrix with ones on the diagonal instead of the Householder matrix. Nevertheless, the motivation of the linIAF differs from ours completely.

The Householder transformations (reflections) were also exploited in the context of learning recurrent neural nets \cite{ASB:16, MJRB:16, WPHLA:16}. They were used for modelling unitary weights instead of more flexible variational posterior, however, these served as a component for representing a matrix of eigenvectors, similarly to our approach.

\section{Experiments}

\subsection{MNIST}

\paragraph{Set-up} In the first experiment we used the MNIST dataset \citep{MNIST} that contains 60,000 training and 10,000 test images of ten handwritten digits ($28 \times 28$ pixels in size). From the training set we put aside 10,000 images for validation and tuning hyper-parameters. We used the dynamically binarized dataset as in \citep{SM:10}.

In order to make more reliable comparison to \citep{RM:15, SKW:15}, we trained the VAE with $40$ stochastic hidden units and the encoder and decoder were parameterized with two-layered neural networks ($300$ hidden units per layer). We used the gating mechanism as the activation function \cite{DG:15, DFAG:16, OKEVGK:16} (see Appendix for details). The length of the HF was $T=\{1, 10\}$ (VAE+HF ($T$=1/10)).\footnote{Taking larger values of $T$ did not resulted in better performance.} For training we utilized ADAM \citep{KB:14} with the mini-batch size equal $100$ and one example for estimating the expected value. The learning rate was set according to the validation set. The maximum number of epochs was $5000$ and early-stopping with a look-ahead of $100$ epochs was applied. We used the \textit{warm-up} \citep{BVVDJB:15, SRMSW:16} for first $200$ epochs. We initialized weights according to \citep{GB:10}.

We compared our approach to linear normalizing flow (VAE+NF) \citep{RM:15}, and finite volume-preserving flows: NICE (VAE+NICE) \citep{DKB:14}, HVI (VAE+HVI) \citep{SKW:15}. When appropriate, the length of a flow $T$ is given. The methods were compared according to the lower bound of marginal log-likelihood measured on the test set.

The code of the proposed approach can be found on: \url{https://github.com/jmtomczak}.

\begin{figure}[!htbp]
\CenterFloatBoxes
\begin{floatrow}
\ttabbox
  {
  \begin{tabular}{ll}
    Method & $\leq \ln p(\mb{x})$ \\
    \hline
    VAE & $-93.89 \pm 0.09$ \\
    \hline
    VAE+HF($T$=1) & $-87.77 \pm 0.05$ \\
    VAE+HF($T$=10) & $-87.68 \pm 0.06$ \\
    \hline
    VAE+NF ($T$=10) \citep{RM:15}	& $-87.5$ \\    
    VAE+NF ($T$=80) \citep{RM:15}	& $-85.1$ \\
	VAE+NICE ($T$=10) \citep{DKB:14}	& $-88.6$ \\
	VAE+NICE ($T$=80) \citep{DKB:14}	& $-87.2$ \\
	VAE+HVI ($T$=1) \citep{SKW:15}	& $-91.70$ \\
	VAE+HVI ($T$=8) \citep{SKW:15}	& $-88.30$
	\end{tabular}
  }
  {\caption{Comparison of the lower bound of marginal log-likelihood measured in nats of the digits in the MNIST test set. For the first three methods the experiment was repeated $3$ times.}\label{tab:mnist}}
\killfloatstyle
\ffigbox
  {\includegraphics[width=0.5\textwidth]{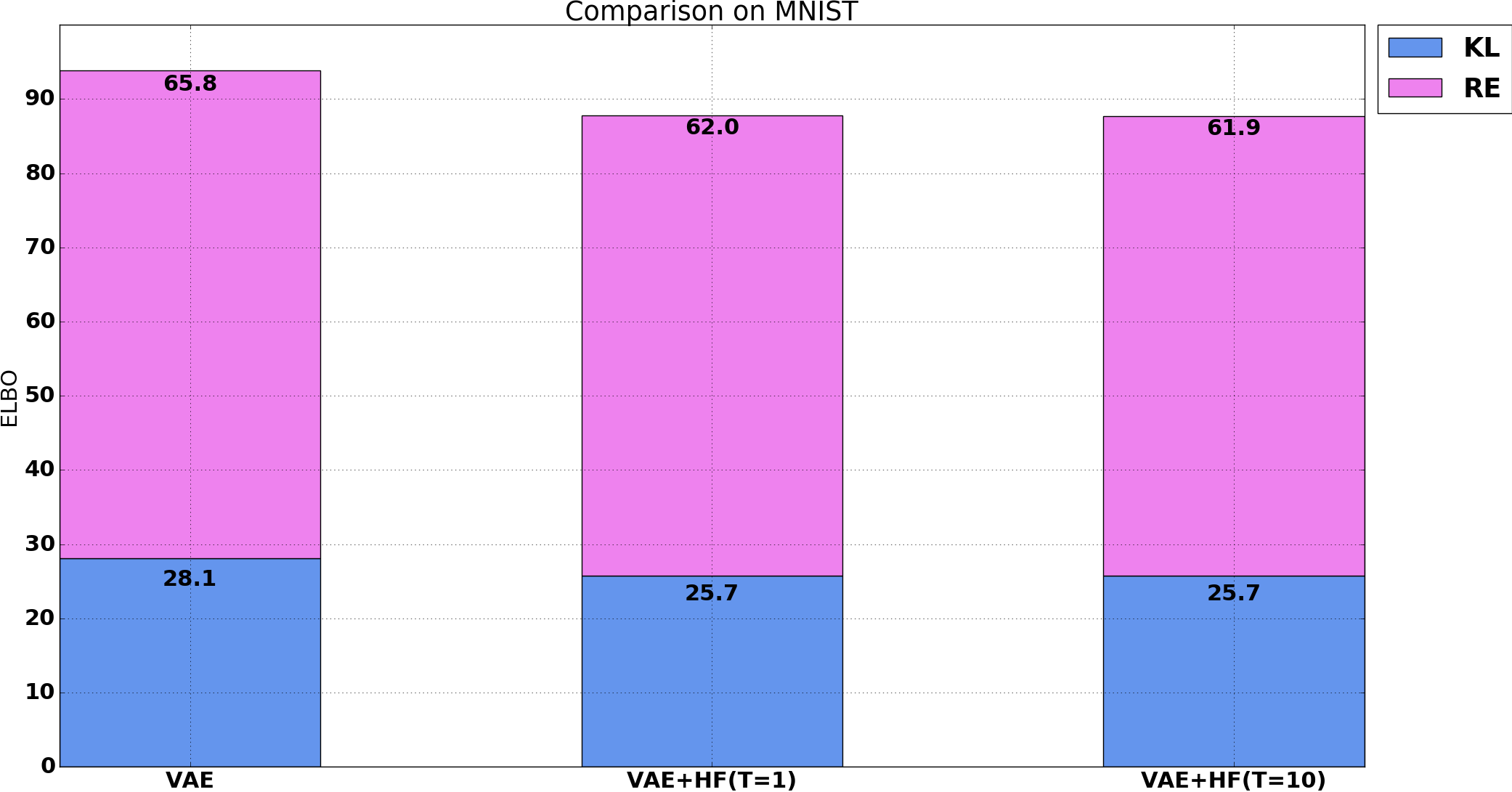}}
  {\caption{Detailed results for VAE, VAE+HF ($T$=1) and VAE+HF ($T$=10). KL denotes the part of ELBO associated with the Kullback-Leibler Divergence and RE correposonds to the reconstruction error (the expected value of the decoder). The numbers represent negative values of KL and RE.}\label{fig:mnist}}
\end{floatrow}
\end{figure}

\paragraph{Discussion} The results presented in Table \ref{tab:mnist} reveal that the proposed flow is competitive comparing to other normalizing flows. This outcome is especially interesting since in our approach sampling from the variational posterior requires application of linear transformations, thus, the HF is reasonably cheap computationally. In our approach we need to store only $T\times M$ more parameters comparing to the vanilla VAE while, \textit{e.g.}, the VAE+NICE and VAE+NF require $\frac{ M \times (M-1)}{2}$ and $\mathcal{O}(T\times M)$ more parameters, respectively. All these remarks are in favor of the proposed flow.

In order to get some insight into the behavior of the HF we also examined the values of components of the training objective (see Figure \ref{fig:mnist}). First, the application of the HF helps the VAE to better model data (lower reconstruction error). Second, the HF gives higher flexibility of the posterior because the approximation of the orthogonal matrix allows the initial posterior to model eigenvalues instead of the whole information about the true posterior, which results in smaller Kullback-Leibler penalty (\textit{i.e.}, the variances do not need to model whole information about the true posterior distribution).

\subsection{Histopathology data}

\begin{figure}[!htbp]
\includegraphics[width=0.4\textwidth]{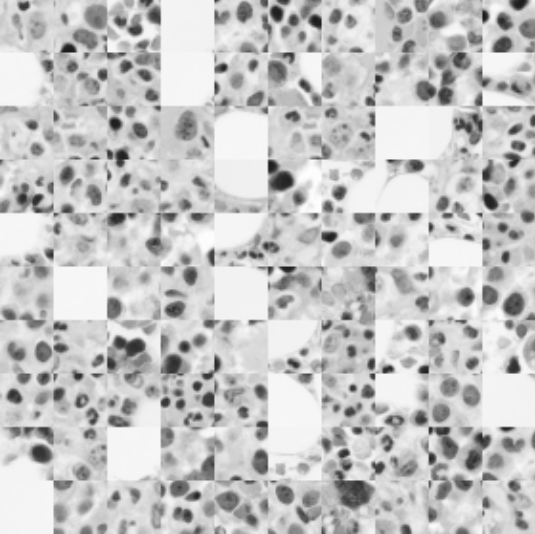}
\vskip -0.25cm
\caption{Selected image patches from the prepared histopathological data.}
\label{fig:histopathology_data}
\end{figure}

\paragraph{Data preparation} In the second experiment we applied the VAE and the VAE+HF to a more challenging problem of grayscale histopathology data. We prepared a dataset basing on histopathological images freely available on-line\footnote{\url{http://www.enjoypath.com/}}. We selected 16 patients\footnote{Patient numbers: 272, 274, 283, 289, 290, 291, 292, 295, 297, 298, 299.} and each histopathological image represented a bone marrow biopsy. Diagnoses of the chosen cases were associated with different kinds of cancer (\textit{e.g.}, lymphoma, leukemia) or anemia. All images were taken using HE, $40\times$, and each image was of size $336 \times 448$. The original RGB representation was transformed to grayscale\footnote{A pixel's value was determined according to the formula: $0.299\times R + 0.587\times G + 0.114\times B$.}. Further, we divided each image into small patches of size $28 \times 28$ (see Figure \ref{fig:histopathology_data} for exemplary image patches). Eventually, we picked 10 patients for training, 3 patients for validation and 3 patients for testing, which resulted in 6,800 training images, 2,000 validation images and 2,000 test images. The selection of patients was performed in such fashion that each dataset contained representative images with different diagnoses and amount of fat.

\begin{wraptable}{r}{0.5\textwidth}
  \caption{Comparison of the lower bound of marginal log-likelihood measured in nats of the image patches in the Histopathology test set. The experiment was repeated $3$ times.}
  \label{tab:histopathology}
  \centering
  \vskip 2mm
  \begin{tabular}{ll}
    Method & $\leq \ln p(\mb{x})$ \\
    \hline
    VAE & $1371.43 \pm 32.12$ \\
    VAE+HF ($T$=1) & $1387.99 \pm 22.09$ \\
    VAE+HF ($T$=10) & $1397.02 \pm 15.15$ \\
    VAE+HF ($T$=20) & $1398.27 \pm 8.11$ \\
	\end{tabular}
\end{wraptable}

\paragraph{Set-up} We used the same architecture of the VAE and the same training procedure as in the previous experiment. Since the data are grayscale, we modeled data using normal distribution instead of the Bernoulli distribution with means constrained to $[0,1]$ by applying the sigmoid function. We compared the vanilla VAE with the VAE+HF ($T$=1,10,20). 

\paragraph{Discussion} The results presented in Table \ref{tab:histopathology} confirmed findings of the previous experiment that the application of the HF helps in obtaining more flexible posterior. Additionally, in this case application of the series of 20 HF steps provided a slight improvement. We have also noticed that the application of the HF allows to obtain less variable results (smaller standard deviation comparing to the vanilla VAE).

\section{Conclusion}

In this paper, we proposed a new volume-preserving flow using Householder transformations. Our idea relies on the observation that the true full-covariance matrix can be decomposed to the diagonal matrix with eigenvalues on the diagonal and the orthogonal matrix of eigenvalues that could be further modeled using a series of Householder transformations. The obtained results in the experimental studies reveal that fully connected VAEs with Householder flows could perform better than other volume-preserving flows. This makes the HF a very promising direction for further investigation. In the future, we aim at using the outlined flow in recently proposed extensions of the variational inference, \textit{e.g.}, importance weighted VAE \citep{BGS:15}, R\'{e}nyi Divergence for VAE \citep{LT:16} and Ladder VAE \citep{SRMSW:16}. Moreover, we believe that the proposed flow can be beneficiary in modelling natural and medical images, where flexibility of the posterior is crucial \citep{OKK:16,SRMSW:16}. We leave investigating this issue for future research.

\subsubsection*{Acknowledgments}
We would like to give special thanks to Diederik P. Kingma for fruitful discussions and insightful remarks on the paper. The research conducted by Jakub M. Tomczak was funded by the European Commission within the Marie Sk\l odowska-Curie Individual Fellowship (Grant No. 702666, ''Deep learning and Bayesian inference for medical imaging'').

%
%
\addcontentsline{toc}{section}{References}

\newpage
\addcontentsline{toc}{section}{Appendix}
\section*{Appendix}
Let $\mb{x}$ be a $D$-dimensional input and $\mb{h}_{1}, \ldots, \mb{h}_{L}$ be hidden units in a neural network (NN).\footnote{To keep the notation uncluttered we set $\mb{h}_{0} = \mb{x}$.} Typically, a hidden layer in NN is calculated using ReLU activation function:
\begin{equation}
\mb{h}_{l} = \max\{0, \mb{W}_{l}\mb{h}_{l-1} + \mb{b}_{l} \},
\end{equation}
where $\mb{W}_{l}, \mb{b}_{l}$ are weights and biases of the $l$-th layer. In fact, ReLU could be expressed using the indicator function $\mathbb{I}[\cdot]$, namely:
\begin{equation}
\mb{h}_{l} = (\mb{W}_{l}\mb{h}_{l-1} + \mb{b}_{l} ) \otimes \mathbb{I}[ \mb{W}_{l}\mb{h}_{l-1} + \mb{b}_{l} > \mathbf{0} ],
\end{equation}
where $\otimes$ denotes the element-wise multiplication.

\begin{figure}[!htbp]
\centering
\includegraphics[width=0.75\textwidth]{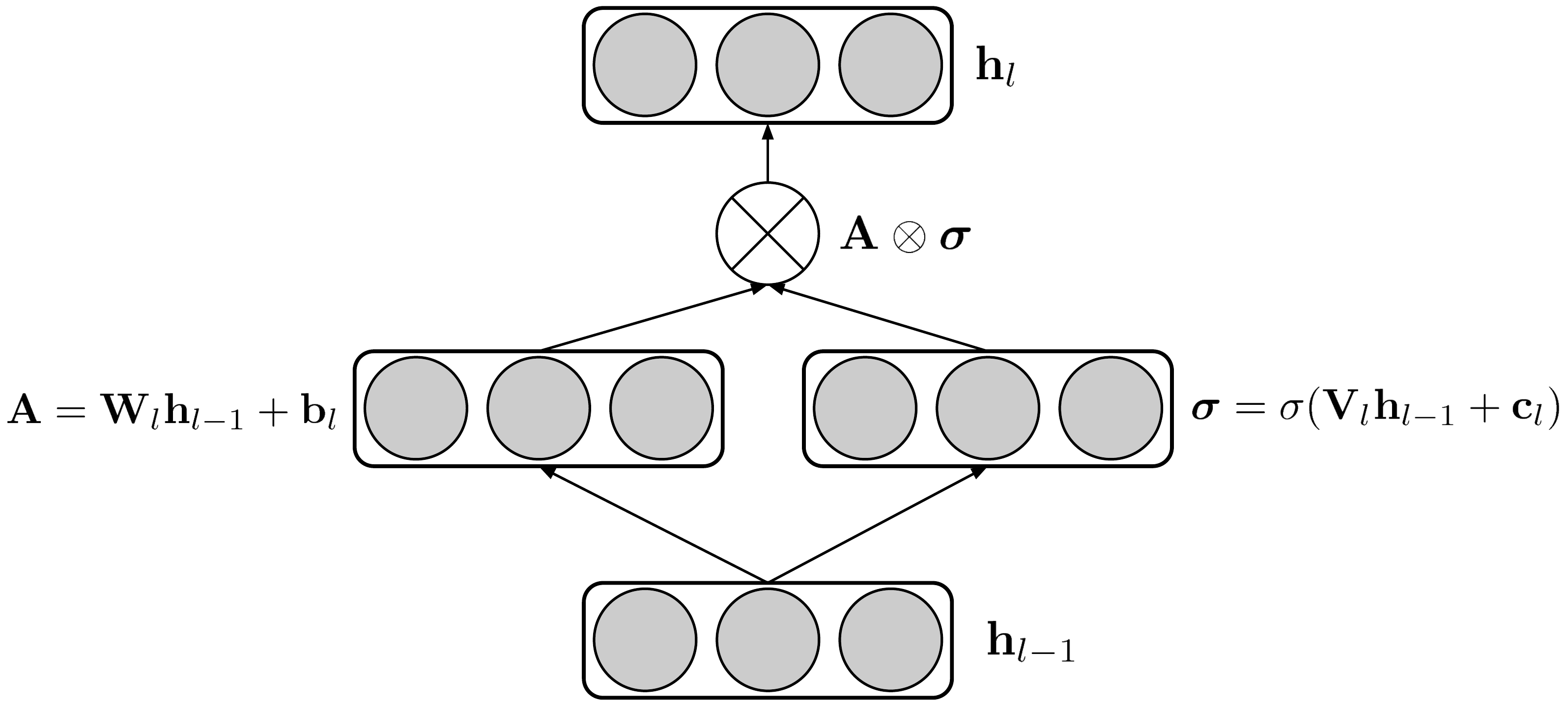}
\caption{A schema of a single layer of the gating mechanism \citep{DFAG:16}.}
\label{fig:gatingMechanism}
\end{figure}

Very recently, a new type of activation function was proposed that applies a gating mechanism \citep{DFAG:16}:
\begin{equation}
\mb{h}_{l} = (\mb{W}_{l}\mb{h}_{l-1} + \mb{b}_{l} ) \otimes \sigma( \mb{V}_{l}\mb{h}_{l-1} + \mb{c}_{l} ),
\end{equation}
Calculation of the gating mechanism for a single layer is presented in Figure \ref{fig:gatingMechanism}.

\end{document}